\documentclass[11pt,a4paper]{article}
\usepackage[hyperref]{acl2020}
\usepackage{times}
\usepackage{latexsym}

\usepackage{graphicx}
\usepackage{comment}
\usepackage{hyperref}
\usepackage{multirow}
\usepackage{multicol}
\usepackage{amsmath}

\usepackage{caption}
\usepackage{subcaption}
\usepackage{stfloats}

\usepackage{booktabs}

\usepackage{microtype}

\aclfinalcopy 

\title{ReINTEL: A Multimodal Data Challenge for \\Responsible Information Identification on Social Network Sites}

\author{Duc-Trong Le$^{1}$,
Xuan-Son Vu$^{2}$,
Nhu-Dung To$^{3}$,
Huu-Quang Nguyen$^{4}$, \\
\textbf{
Thuy-Trinh Nguyen$^{4}$, 
Linh Le$^{4}$,
Anh-Tuan Nguyen$^{4}$,
Minh-Duc Hoang$^{4}$,
Nghia Le$^{4}$} \\
\textbf{
Huyen Nguyen$^{5}$,
Hoang D. Nguyen$^{6}$} \\\\
$^{1}$University of Engineering and Technology, Vietnam National University, Vietnam. \\ \texttt{trongld@vnu.edu.vn}\\
$^{2}$Dept. of Computing Science, Ume\r{a} University,
Sweden. \\ \texttt{sonvx@cs.umu.se}\\
$^{3}$School of Computer Science, University of Sydney,
Australia. \\ \texttt{duto3894@uni.sydney.edu.au}\\
$^{4}$ReML.AI - Reliable Machine Learning Lab, International. \\\texttt{\{quang, trinh, linh, tuan, duc, nghia\}@reml.ai} \\
$^{5}$Hanoi University of Science, Vietnam National University, Vietnam.\\ \texttt{huyenntm@hus.edu.vn}\\
$^{6}$School of Computing Science, University of Glasgow, 
Singapore. \\\texttt{harry.nguyen@glasgow.ac.uk}
\\}

\date{}

\begin{document}
\maketitle

\begin{abstract}

This paper reports on the ReINTEL Shared Task for Responsible Information Identification on social network sites, which is hosted at the seventh annual workshop on Vietnamese Language and Speech Processing (VLSP 2020). Given a piece of news with respective textual, visual content and metadata, participants are required to classify whether the news is \textit{`reliable'} or \textit{`unreliable'}. In order to generate a fair benchmark, we introduce a novel human-annotated dataset of over 10,000 news collected from a social network in Vietnam. All models will be evaluated in terms of AUC-ROC score, a typical evaluation metric for classification. The competition was run on the Codalab platform. Within two months, the challenge has attracted over 60 participants and recorded nearly 1,000 submission entries.

\end{abstract}

\section{Introduction}
This challenge aims at identifying the reliability of information shared on social network sites (SNSs). With the blazing-fast spurt of SNSs (e.g. Facebook, Zalo and Lotus), there are approximately 65 million Vietnamese users on board with the annual growth of 2.7 million in the recent year, as reported by the Digital 2020 \footnote{\url{https://wearesocial.com/digital-2020}}. SNSs have become widely accessible for users to not only connect friends but also freely create and share diverse content \cite{shu2017fake,zhou2019fake}. A number of users, however, has exploited these social platforms to distribute fake news and unreliable information to fulfill their personal or political purposes (e.g. US election 2016 \cite{allcott2017social}). It is not easy for other ordinary users to realize the unreliability, hence, they keep spreading the fake content to their friends. The problem becomes more seriously once the unreliable post becomes popular and gains belief among the community. Therefore, it raises an urgent need for detecting whether a piece of news on SNSs is reliable or not. This task has gained significant attention recently~\cite{CSI:2017,shu2019defend,shu2019beyond,yang2019unsupervised}. 

The shared task focuses on the responsible (i.e. reliable) information identification on Vietnamese SNSs, referred to as ReINTEL. It is a part of the 7th annual workshop on Vietnamese Language and Speech Processing, VLSP 2020\footnote{\url{https://vlsp.org.vn/vlsp2020}} for short. As a binary classification task, participants are required to propose models to determine the reliability of SNS posts based on their content, image and metadata information (e.g. number of likes, shares, and comments). The shared task consists of three phases namely \textit{Warm up, Public Test,  Private Test}, which is hosted on Codalab from October 21st, 2020 to November 30th, 2020. In summary, there are around 1000 submissions created by 8 teams and over 60 participants during the challenge period. 

As our first contribution, this shared task provides an evaluation framework for the reliable information detection task, where participants could leverage and compare their innovative models on the same dataset. Their knowledge contribution may help improve safety on online social platforms. Another valuable contribution is the introduction of a novel dataset for the reliable information detection task. The dataset is built based on a fair human annotation of over 10,000 news from SNSs in Vietnam. We hope this dataset will be a useful benchmark for further research. In this shared task, AUC-ROC is utilized as the primary evaluation metric. 

The remainder of the paper is organized as follows. The next section describes the data collection and annotation methodologies. Subsequently, the shared task description and evaluation are summarized in Section 3. In Section 4, we discusses the potentials of language and vision transfer learning for the detection task. Section 5 describes the competition, approaches and respective results. Finally, Section 6 concludes the paper by suggesting potential applications for future studies and challenges.

\section{The ReINTEL 2020 Dataset}
\subsection{Data Collection}

\begin{figure*}
	\centering
	\includegraphics[width=.97\linewidth,height=7.0cm]{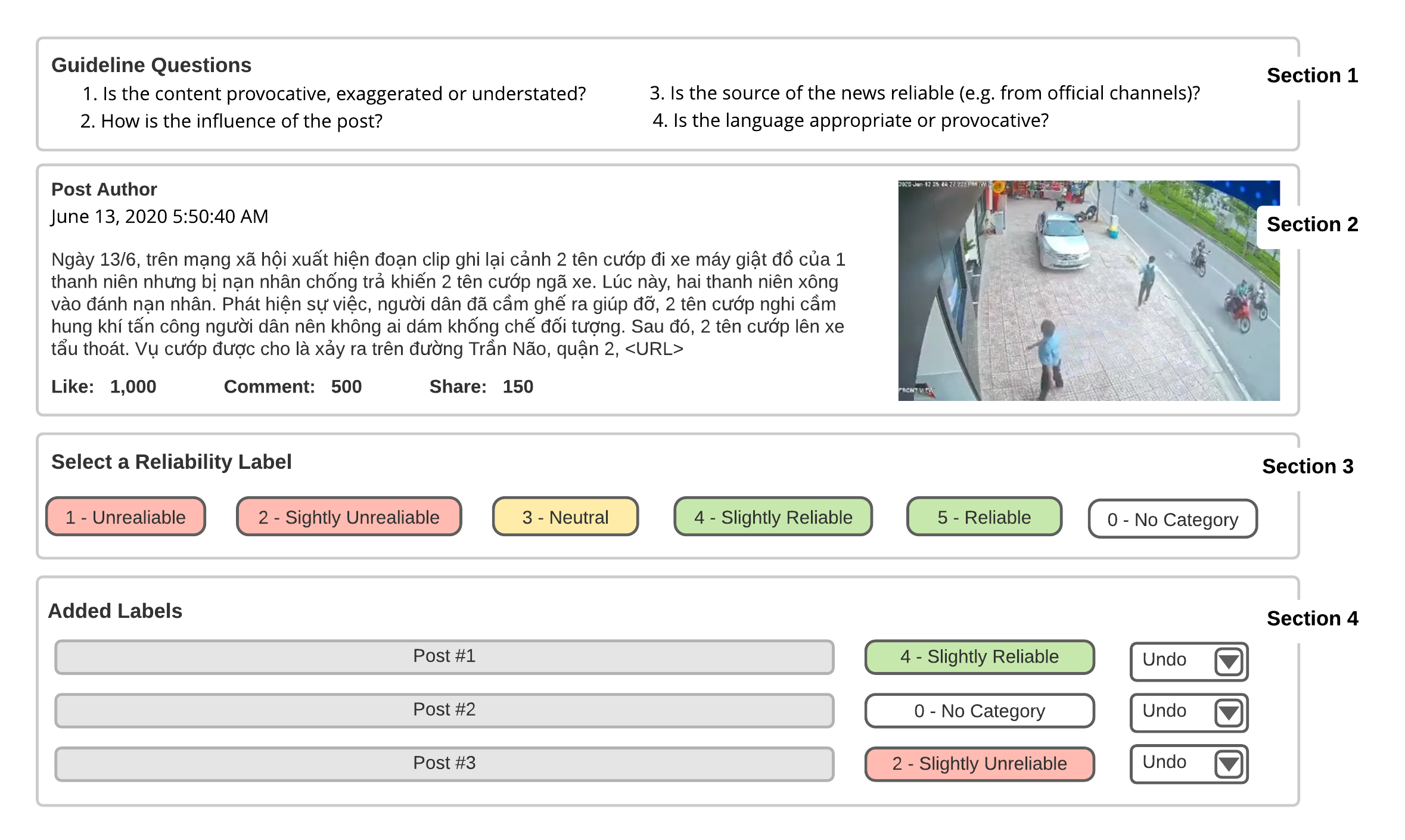}
	\caption{Data Annotation Tool}
	\label{fig:reintel_anno_tool}
\end{figure*}

We collect the data for two months from August to October 2020. There are two main sources of the data: SNSs and Vietnamese newspapers. As for the former source, public social media posts are retrieved from news groups and key opinion leaders (KOLs). Many fake news, however, has been flagged and removed from the social networking sites since the enforcement of Vietnamese cybersecurity law in 2019 \cite{tuanson_2018}. Therefore, to include the deleted fake news, we gather newspaper articles reporting these posts and recreate their content.

All the collected data were originally posted in the period of March - June 2020. During this time, Vietnam was facing a second wave of Covid-19 with a drastic increase from 20 to 355 cases \cite{who}. The spread of Covid-19 results in an ‘infodemic’ in which misleading information is disseminated rapidly especially on social media \cite{hou2020assessment, huynh2020covid}. Hence, this period is a potential source of fake news. Besides Covid-19, the items in our dataset cover a wide range of domains including entertainment, sport, finance and healthcare. The result of the data collection stage is 10,007 items that are prepared for the annotation process.

\subsection{Data Annotation}
\label{dataset}
\subsubsection{Annotator and Training}
We recruit 23 human annotators to participate in the annotation process. The annotators receive one week training to identify fact-related posts and how to evaluate the reliability of the post based on primary features including the news source, its image and content. 

\subsubsection{Annotation Tool}
Figure \ref{fig:reintel_anno_tool} demonstrates the annotation tool interface, which is designed to support quick and easy annotation. The first section contains guideline questions to remind the annotators of the labeling criterion including the news source credibility, the language appropriateness and fact accuracy. 
The second section is the post content, image and influence (i.e. number of likes, comments and shares). In Section 3, the annotators select a Reliability score for the post. There is a 5-point reliability Likert scale for fact-based posts with the following labels: 1 - Unreliable, 2 - Slightly unreliable, 3 - Neutral, 4 - Slightly reliable, 5 - Reliable. On the other hand, if the post is opinion-based and does not contain facts, the annotators should select label ‘0 - No category’ instead.

The last section is a list of labeled items for the annotators to review and update their decision, if necessary, using the ‘Undo’ button.

\subsubsection{Annotation Process}
The annotation process is conducted from 9th to 19th October 2020. The annotators are divided into three groups to annotate 10,007 items independently. Therefore, each item will be annotated three times by different annotators. 

Once the annotators finish 30,021 annotations (i.e. 10,007 items annotated three times), we filter and summarise the result based on majority vote basis. Firstly, we combine labels of the same essence: Category 1 and 2 (Unreliable and Sightly unreliable) and Category 4 and 5 (Slightly reliable and Reliable). After merging the categories, we select the majority votes to be the final labels.  If the majority vote is 1 or 2, the final label should be 1 - Unreliable. If the majority vote is 4 or 5, the final label should be 0 - Reliable. When the majority vote is 3 - Neutral, we finalise using ground truth labels. Lastly, if the majority agrees that the post is not fact-based (i.e. 0 - No Category), we remove it from the set.

\begin{figure*}
	\centering
	\includegraphics[width=.99\linewidth,height=6.5cm]{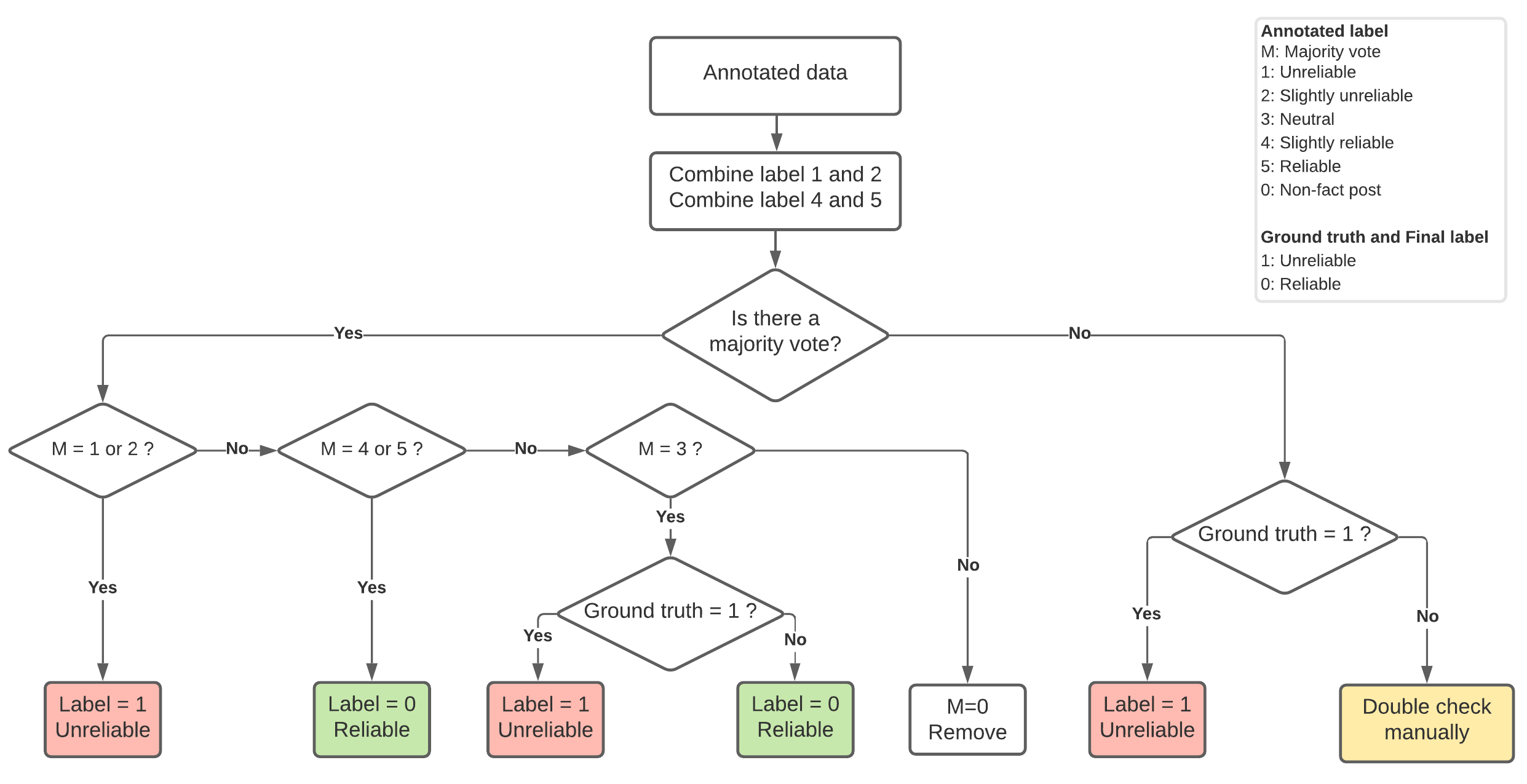}
	\caption{Data Annotation Process}
	\label{fig:reintel_anno_process}
\end{figure*}

For items with no majority votes (i.e. three annotators have different opinions), we follow an alternate procedure. If the ground truth label is 1 - unreliable, the final label should be 1. On the other hand, if the ground truth label is 0 - reliable, we double check to separate reliable news from opinion-based items. The process is illustrated in Figure \ref{fig:reintel_anno_process}.


\subsubsection{Content Filtering}

Once the annotation process is finished, data needs to go through the last step before being published for the competition – the content filtering. In this step, we manually check to ensure that data, including both text and image, published for the competition:

\begin{enumerate}
    \item Does not violate any law, statue, ordinance, or regulation
    \item Will not give rise to any claims of invasion of privacy or publicity
    \item Does not contain, depict, include or involve any of the following:
    \begin{itemize}
        \item Political or religious views or other such ideologies
        \item Explicit or graphic sexual activity
        \item Vulgar or offensive language and/or symbols or content
        \item Personal information of individuals such as names, telephone numbers, and addresses
        \item Other forms of ethical violations
    \end{itemize}
\end{enumerate}





\section{The ReINTEL 2020 Challenge}

\subsection{Dataset Splitting}
Data splitting for data challenge is a difficult process in order to avoid evidence ambiguity and concept drifting which are the main cause of unstable ranking issue in data challenges.

In this competition, we apply RDS~\cite{nguyen2020reinforced} to split ReINTEL data into three sets including public train, validation, and private test sets. It is worth to mention that, RDS is a method to approximate optimum sampling for model diversification with ensemble rewarding to attain maximal machine learning potentials. It has a novel stochastic choice rewarding is developed as a viable mechanism for injecting model diversity in reinforcement learning. 

\subsubsection{Baselines}
To apply RDS~\cite{nguyen2020reinforced} for the data splitting process, it requires to have baseline learners to obtain rewards for the reinforced process. It is recommended to choose representative baseline learners, to let the reinforced learner better capture different learning behaviors. The use of these baseline learners is important since each learner will behave differently depending on the patterns contained in the target data. As a result, RDS helps to increase the diversity of the data samples in different sets. Here we employ three models to classify reliable news using textual features as follows:

\begin{itemize}
    \item \textbf{Bi-LSTM}~\cite{Schuster:bi-lstm} is a bi-directional LSTM model. It has two LSTMs in which, one LSTM takes input sequence in a forward direction, and another LSTM takes input sequence in a backward direction. The use of Bi-LSTM architecture helps to increase the amount of information available to the network, to gain better performance in most of sequence related tasks. Bi-LSTM network is a standard baseline for most of text classification tasks.
    \item \textbf{CNN-Text}~\cite{kim-2014-convolutional} is the use of CNN~\cite{CNN:1989} network on word embeddings to perform the classification tasks. The simple architecture outperformed all other models at the publication time. 
    \item \textbf{EasyEnsemble}~\cite{EasyEnsemble:2009} is used to represent a tradition approach in dealing with im-balanced dataset. For the vectorization, we trained a Sent2Vec~\cite{pgj2017unsup} using the combined 1GB texts of Vietnamese Wikipedia data~\cite{vu:2019n} and 19 GB texts of~\newcite{newscorpus:2018}.
\end{itemize}

\subsubsection{Learning Dynamics}

\begin{figure}[h]
    \centering
    \begin{subfigure}[b]{0.2\textwidth}
        \centering
        \includegraphics[width=\textwidth]{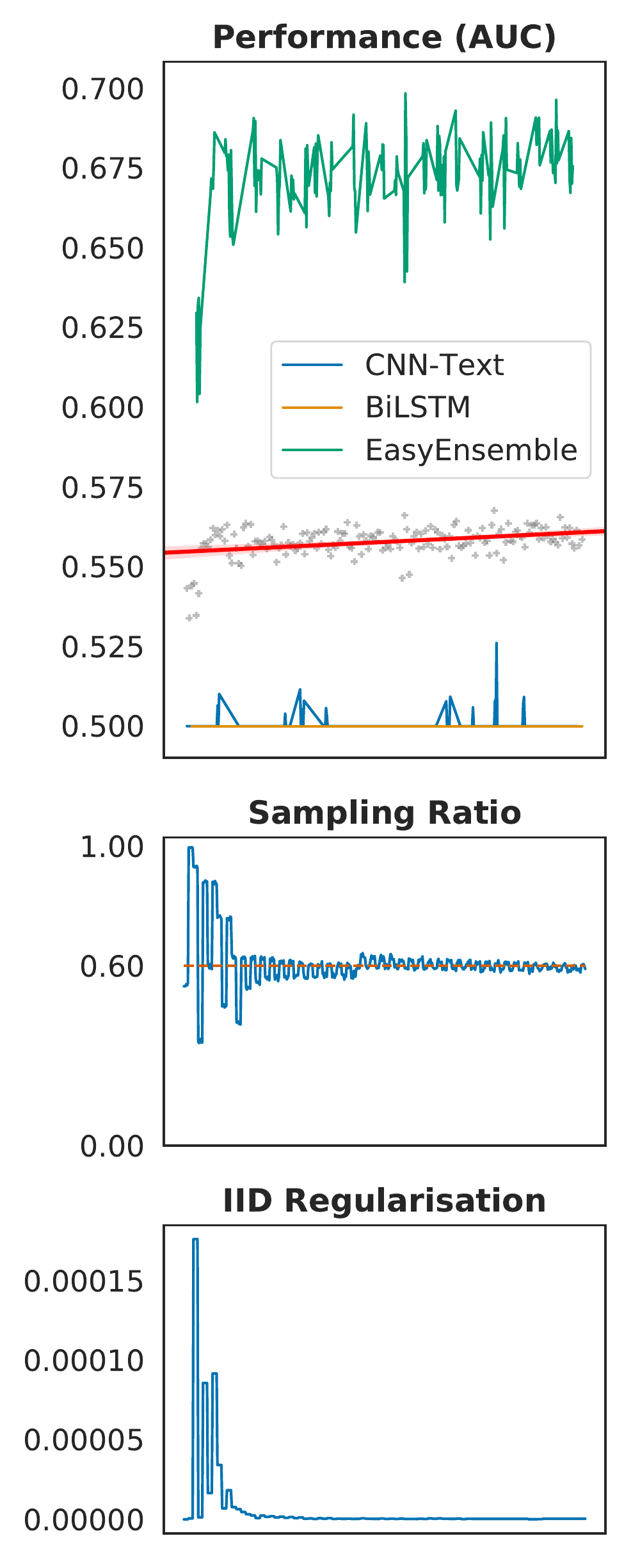}
        \caption{Training Set}
        \label{fig:reintel_sto_trainrest}
    \end{subfigure}
    \begin{subfigure}[b]{0.2\textwidth}
        \centering
        \includegraphics[width=\textwidth]{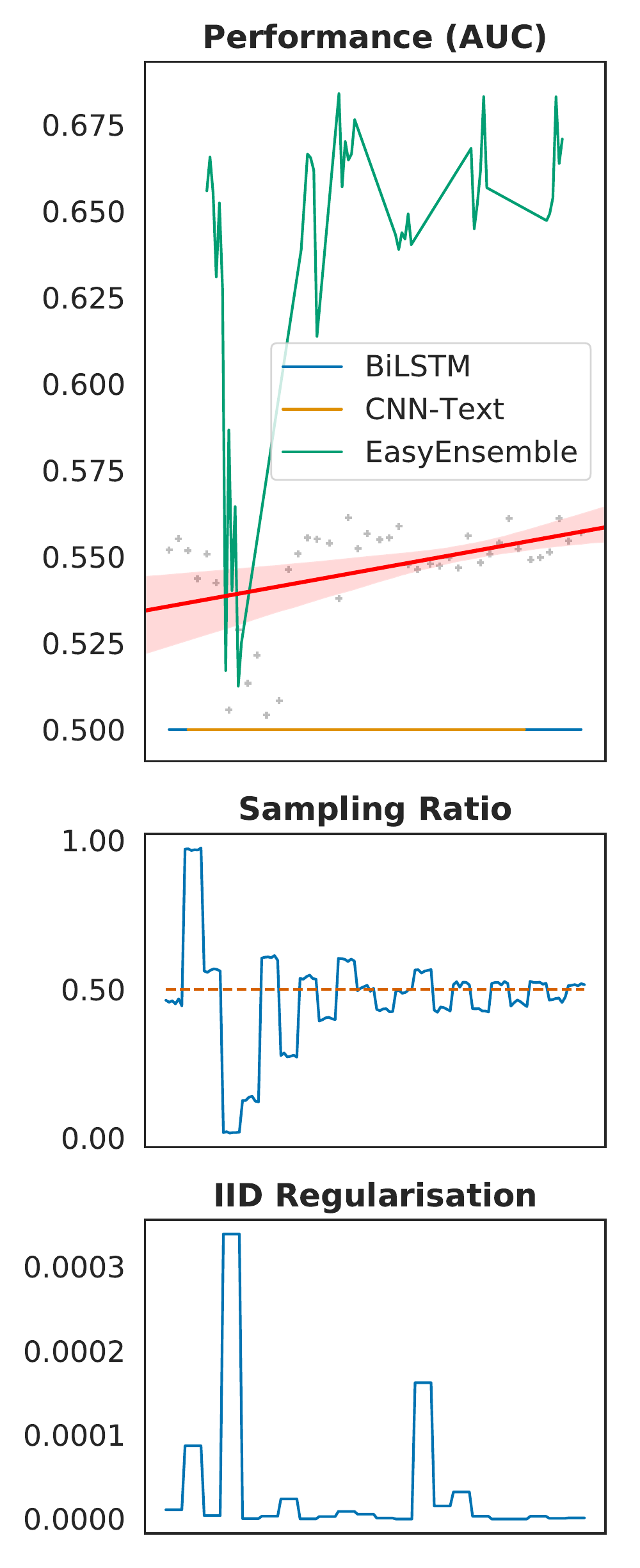}
        \caption{Val \& Test Sets}
        \label{fig:reintel_sto_valtest}
    \end{subfigure}
    \caption{Learning Dynamics for splitting data into 3 sets (public training, public testing, and private testing) using RDS Stochastic Choice Reward Mechanism~\cite{nguyen2020reinforced}.}
    \label{fig:learning_sto}
\end{figure}

To disentangle dataset shift and evidence ambiguity of the data splitting strategy, we apply RDS stochastic choice reward mechanism \cite{nguyen2020reinforced} to create public training, public- and private testing sets. Figure~\ref{fig:learning_sto} illustrates the learning dynamic towards the goal. 

\section{Transfer Learning}
Knowledge transfer has been found to be essential when it comes to downstream tasks with new datasets. If this transfer process is done correctly, it would greatly improve the performance of learning. Since ReINTEL challenge is a multimodal challenge, both visual based knowledge transfer and language based knowledge transfer are used by different teams. 

To be fair between participants, we required all teams to register for the use of pre-trained models. 
 \autoref{tbl:pretrained_models} lists all pre-trained language and vision models registered by all participants. 

\subsection{Language Transfer Learning}
For natural language processing tasks in Vietnamese, there have been many pre-trained language models are available. In 2016, ~\newcite{word2vecvn_2016} introduced the first monolingual pre-trained models for Vietnamese based on Word2Vec~\cite{mikolov2013efficient}. The use of pre-trained Word2VecVN models was proved to be useful in various tasks, such as the name entity recognition task~\cite{VU:2018}. In 2019, ~\newcite{vu:2019n} introduced the use of multiple pre-trained language models to achieve new state-of-the-art results in the name entity recognition task~\cite{Nguyen:19}. Up to date, there have been many other new monolingual language models for Vietnamese are available such as PhoBERT~\cite{phobert}, vElectra and ViBERT~\cite{the2020improving}.


\begin{table*}[]
\begin{tabular}{p{6cm}|l|l|p{5.7cm}}
\toprule
Model                  & Language & Vision & Description \\ \midrule
Word2VecVN~\cite{word2vecvn_2016}  & x   &        & Trained on 7GB texts of Vietnamese news        \\ \hline
FastText (Vietnamese version)~\cite{joulin2016fasttext}    & x   &        &  Trained on Vietnamese texts of the CommonCrawl corpus  \\ \hline
ETNLP~\cite{vu:2019n}                  & x   &        & Trained on 1GB texts of Vietnamese Wikipedia     \\ \hline
PhoBERT~\cite{phobert} & x   &        &  Trained on 20GB texts of both Vietnamese news and Vietnamese Wikipedia \\ \hline
Bert4News~\cite{bert4news}    & x   &        &  Trained on more than 20GB texts of Vietnamese news  \\ \hline
vElectra and ViBERT~\cite{the2020improving}   & x   &       & vElectra was trained on 10GB texts, whereas ViBERT was trained on 60GB texts of Vietnamese news            \\ \hline
VGG16~\cite{simonyan2015deep}                  &     & x      & Trained on ImageNet~\cite{imagenet_cvpr09}            \\ \hline
YOLO~\cite{yolo:2018}    &     & x      &        Trained on ImageNet~\cite{imagenet_cvpr09}     \\ \hline
EfficientNet B7~\cite{EfficientNet:2019}    &     & x      &    Trained on ImageNet~\cite{imagenet_cvpr09}          \\

\bottomrule
\end{tabular}
\caption{List of pre-trained models registered by all participants of ReINTEL challenge in 2020.}
\label{tbl:pretrained_models}
\end{table*}

\subsection{Vision Transfer Learning}
Different from language models, visual models are normally universal and existing pre-trained models can be directly applied in most of image processing tasks. For the use of visual features, there is only one team using multimodal features among top 6 teams of the leader board. This team, in fact, achieved the $1^{st}$ rank on the public test (see \autoref{table2:approaches}); but they did not get the same rank on the private test. This hints that the reliability of news mainly depends on content of news and other meta information, such as number of likes on social networks. Moreover, it is yet to be explored to capture the reliability of news using both vision and language information.

\subsection{Language and Vision Transfer Learning}
The use of both language and vision transfer learning is important for multimodal tasks. This line of research has attracted much attention with various new language-vision models, such as VilBERT~\cite{vilbert:2019}, 12-in-1~\cite{12in1_cvpr:2020}. 
No participants employ into this approach in the ReINTEL challenge due to the lack of language and vision pre-trained models in Vietnamese. Moreover, it is required to have extensive computer resources for applying this approach in a data challenge. In the future, we expect to see more research done in this direction because both images and texts are essential to SNS issues.
\section{Results}

\subsection{Data Format}
Each instance includes 8 main attributes with/without a binary target label. Table \ref{table: data_attribute} summarizes the key features of each attribute.

\begin{table*}[ht!] 
\begin{tabular}{@{}llp{11.5cm}l@{}}
\toprule
\textbf{No} & \textbf{Attribute} & \textbf{Description}             \\ 
\midrule
1  & id                 & Unique ID of each post                    \\
2  & user\_name         & Anomynized post owner’s identity          \\
3  & post\_message      & Text content of the post                  \\
4  & timestamp\_post    & The time when the post is uploaded        \\                                                            
5  & num\_like\_post    & Number of likes that the post received    \\
6  & num\_comment\_post & Number of comments that the post received \\                                                
7  & num\_share\_post   & Number of shares that the post received   \\
8  & image              & The image uploaded with the post          \\
9  & label              & \begin{tabular}[c]{@{}l@{}}Manually annotated label indicating the reliability of the post\\ 1: Unreliable\\ 0: Reliable\end{tabular}   \\ \bottomrule
\end{tabular}
\caption{Data attributes}
\label{table: data_attribute}
\end{table*}

\subsection{Training/Testing Data}

The challenge provides approximately 8,000 training examples with the respective target labels. The testing set consists of 2,000 examples without labels. 

\subsection{Result Submission} 

Participants must submit the result in the same order as the testing set in the following format: 
\begin{verbatim}
    id1, label probability 1
    Id2, label probability 2
    …
\end{verbatim}

\subsection{Evaluation Metric}
The challenge task is evaluated based on Area Under the Receiver Operating Characteristic Curve (AUC-ROC), which is a typical metric for classification tasks. Let us denote $X$ as a \textit{continuous random variable} that measures the `classification' score of a given a news. As a binary classification task, this news could be classified as \textit{"unreliable"} if $X$ is greater than a threshold parameter $T$, and \textit{"reliable"} otherwise. We denote $f_1(x), f_0(x)$ as probability density functions that the news belongs to \textit{"unreliable"} and \textit{"reliable"} respectively, hence the true positive rate $TPR(T)$ and the false positive rate $FPR(T)$ are computed as follows:
\begin{align}
    TPR(T) &= \int_{T}^{\infty}f_1(x)dx \\
    FPR(T) &= \int_{T}^{\infty}f_0(x)dx
\end{align} and the AUC-ROC score is computed as:
\begin{align}
    AUC\_ROC &= \int_{-\infty}^{\infty}TPR(T)FPR'(T)dT
\end{align}
Here, submissions are evaluated with ground-truth labels using the \textit{scikit-learn}'s implementation \footnote{\url{https://scikit-learn.org/stable/modules/generated/sklearn.metrics.roc_auc_score.html}}.

\begin{table*}[h!]
\caption{Top 6 teams on public-test and private-test with submitted papers and their final approaches. The rank is based on the ROC-AUC scores on the private-test.}
\centering
\label{table2:approaches}
\scalebox{0.85}{
\begin{tabular}{l|p{1.8cm}|c|c|p{6.6cm}|l|l}
\toprule
\multirow{2}{*}{\#} & \multirow{2}{*}{Team} & \multicolumn{2}{c|}{ROC-AUC} & \multirow{2}{*}{Final Approach}  & \multirow{2}{*}{Ensemble?} & \multirow{2}{*}{Multimodal?}       \\ \cline{3-4}
& & Public-test & Private-test &   & &      \\ \hline
1 & Kurtosis & 0.9399 & \textbf{0.9521} & TF-IDF + SVD; Emb + SVD; NB, LightGBM, CatBoost & Yes & No\\ \hline
2 & NLP\_BK & 0.9360 & 0.9513 &  Bert4News + phoBERT + XLM + MetaFeatures & Yes & No \\ \hline
3 & SunBear & {0.9418} & 0.9462 & RoBerta + MLP & Yes & No \\ \hline
4 & uit\_kt  & - & 0.9452 &    phoBERT + Bert4News & Yes & No\\ \hline
5 & Toyo-Aime & \textbf{0.9427} & 0.9449  & CNN + Bert + Fully connected & Yes & Yes          \\ \hline
6 & ZaloTeam & - & 0.9378  &viBERT + viELECTRA + phoBERT & Yes & No          \\ \bottomrule
\end{tabular}
}
\end{table*}

\subsection{Participation}
During the course two months of the competition, 61 participants sign up for the challenge. 30\% of the participants compete in groups of 2 (6 teams) and 4 members (2 teams). 19 participants sign our corpus usages agreement. 

From top 8 of the Private test leaderboard, 6 teams/participants submit their technical reports that demonstrate their strategies and findings from the challenge. The summary of the competition participation can be seen in Table \ref{table: participation_summary}.

\begin{table}[h!]
\begin{tabular}{@{}lp{2.6cm}@{}}
\toprule
\textbf{Metric}             &  Value  \\ \midrule
Number of participants      & 61 \\
Number of teams             & 8  \\
Number of signed agreements & 19 \\
Number of submitted papers  & 6  \\ \bottomrule
\end{tabular}
\caption{Participation summary}
\label{table: participation_summary}
\end{table}

\subsection{Outcomes}

In total, 657 successful entries were recorded. The highest results of the Public test and Private test phase were 0.9427 and 0.9521 respectively. Key descriptive statistics of the results in each phase is illustrated in Table \ref{table: results_summary}.

\begin{table}[h!]
\resizebox{\columnwidth}{!}{%
\begin{tabular}{@{}llll@{}}
\toprule
              & \textbf{Public Test} & \textbf{Private Test} & \textbf{Overall} \\ \midrule
Total Entries & 571                  & 86                    & 657              \\
Highest ROC   & 0.9427               & 0.9521                & 0.9474           \\
Mean ROC      & 0.8463               & 0.8942                & 0.8703           \\
Std. ROC      & 0.1215               & 0.1022                & 0.1119          \\ \bottomrule
\end{tabular}
}
\caption{Results summary}
\label{table: results_summary}
\end{table}

\section{Conclusion}

The rise of misleading information on social media platforms has triggered the need for fact-checking and fake news detection. Therefore, the reliability of news has become a critical question in the modern age. In this paper, we introduce a novel dataset of nearly 10,000 SNSs entries with reliability labels. The dataset covers a great variety of topics ranging from healthcare to entertainment and economics. The annotation and validation process are presented in details with several filtering rounds. With both linguistic and visual features, we believe that the corpus is suitable for future research on fake news detection and news distributor behaviours using NLP and computer vision techniques. In Vietnam, where datasets on SNSs are scarce, our corpus will serve as a reliable material for other research.

\section*{Acknowledgment}
The authors would like to thank the InfoRE company for the data contribution, the ReML-AI research group\footnote{\url{https://reml.ai}} for the data contribution and financial support, and the twenty three annotators for their hard work to support the shared task. Without their support, the task would not have been possible.

\bibliography{acl2020}
\bibliographystyle{acl_natbib}
\end{document}